# Machine Learning Assisted Inverse Design of Microresonators

ARGHADEEP PAL,[1,2] ALEKHYA GHOSH,[1,2] SHUANGYOU ZHANG[1], TOBY BI,[1,2] AND PASCAL DEL'HAYE[1,2,*]

[1]*Max Planck Institute for the Science of Light, 91058 Erlangen, Germany*
[2]*Department of Physics, Friedrich-Alexander-Universität Erlangen-Nürnberg, 91058 Erlangen,Germany*
*\*pascal.delhaye@mpl.mpg.de*

**Abstract:** The high demand for fabricating microresonators with desired optical properties has led to various techniques to optimize geometries, mode structures, nonlinearities and dispersion. Depending on applications, the dispersion in such resonators counters their optical nonlinearities and influences the intracavity optical dynamics. In this paper, we demonstrate the use of a machine learning (ML) algorithm as a tool to determine the geometry of microresonators from their dispersion profiles. The training dataset with ~460 samples is generated by finite element simulations and the model is experimentally verified using integrated silicon nitride microresonators. Two ML algorithms are compared along with suitable hyperparameter tuning, out of which Random Forest (RF) yields the best results. The average error on the simulated data is well below 15%.

## 1. Introduction

The small mode volumes and high quality (Q) factors of microresonators have made them excellent tools to confine light and steer up high nonlinear effects with low input power threshold [1]. One example is four-wave mixing (FWM), which leads to the formation of optical frequency combs [2]. These combs have shown great applications [3], mainly in the fields of spectroscopy [4], sensing [5], telecommunications [6], and waveform generation [7], due to their broad spectrum with uniform spacings. A sufficiently good overlap of the equidistant FWM generated sidebands and the respective resonance frequencies of the microresonator is needed to generate broadband frequency combs. Thus, the comb formation is strongly influenced by dispersion, which leads to uneven spacings between the resonance frequencies. These deviations in resonances from the equidistant positions are evaluated via the integrated dispersion $D_{int}$, which can be written as

$$D_{\text{int}}(m) = \omega_m - \omega_0 - D_1 \times (m - m_0). \tag{1}$$

Here, $\omega_m$ is the angular resonance frequency of the $m^{th}$ mode with respect to the pump mode $m_0$ at angular pump frequency $\omega_0$. $D_1/2\pi$ is the free spectral range (FSR) around the pump mode. Resonator dispersion also plays an important role in the temporal intracavity soliton dynamics. Out of many applications, the generation of bright Kerr frequency combs occurs in presence of anomalous dispersion [8], whereas, normal dispersion [9] favors the formation of dark pulse solitons as shown in Fig. 1(a), where $D_{\text{int}}$ has been plotted as a function of wavelength for $Si_3N_4$ ring resonators with similar radii but different core heights and widths. Recently, it has been demonstrated that dark-bright soliton bound states can be generated in a microresonator crossing different dispersion regimes [10], leading to light states with close to constant output power but resembling a comb in frequency domain. In addition, resonators with zero dispersion not only exhibit different soliton dynamics, but also are desirable for ultrabroadband comb generation [11,12]. Therefore, proper engineering of the dispersion becomes vital [13,14]. Dispersion calculations can be carried out via simulations and experiments. A quick analysis of dispersion is not possible by the conventional simulation techniques due to their iterative nature leading to high time-complexity. Most importantly, this simulation or measurement is a unidirectional process, i.e., from the structure, one can get an idea about the dispersion but not the other way around. Methods of the inverse design of microresonators [15,16] usually follow the approach of starting with a structure from empirical knowledge and then iteratively modifying the geometries based on the deviation from the desired dispersion profile. Owing to the highly nonlinear relation between the geometrical specifications of a microresonator structure and its dispersion profile, a simple method for structure prediction is not feasible. Advanced optimization methods can help us in that aspect, however, extensive numerical simulations are required in each step of the process, making the process slow and time-consuming.

ML is highly successful to find inherently complex nonlinear relationships between datasets in domains like, signal processing [17], natural language processing [18], computer vision [19] and data mining [20]. Due to the vast success, applications of ML transcend the boundaries of computer science to fields like classical and quantum photonics as well [21,22]. Propagation of pulses through a nonlinear medium has been effectively modelled by a recurrent neural network (RNN) [23]. ML algorithms have been utilized for predicting dispersion relations of photonic crystals [24], estimating effective refractive index, mode area, dispersion and other parameters of photonic crystal fibers [25], as well as for developing photonic crystal based gas sensors [26]. In the problem of photonic inverse designing, various ML approaches show impressive efficiencies [27]. A deep learning (DL) based bidirectional model has successfully predicted subwavelength geometry of nanoparticles from far-field optical response [28] and also discusses advantages of using ML methods over evolutionary algorithms. For a desired scattering profile, geometries of the nanoparticles have been predicted using artificial neural network (ANN) [29]. A generative adversarial network (GAN) based model has been used to predict metasurface geometries [30].

In this work, we propose a novel method to predict the structure of the microresonators based on the dispersion parameter $D_{\text{int}}$ using ML. For the inverse design, a regression ML model has been trained by creating a dataset simulating dispersion variation over wavelength for different microresonator geometries. The simulations have been carried out using the finite element method (FEM) in COMSOL. After the validation of the model with the test data, this ML technique is used to estimate the heights, widths, and radii of the fabricated resonators from the experimentally obtained dispersion plots. Among different dispersion measurement techniques available [13], we chose frequency comb assisted diode laser spectroscopy [31].

## 2. Experimental procedures and simulations

The dispersion simulations are performed with structural and material parameters as input as shown in Fig. 1(b). To initiate the FEM solver, an approximate resonance frequency is needed which is given by

$$f_{\text{guess}} = \frac{mc}{2\pi R n}, \tag{2}$$

where $c$ is the speed of light, $m$ is the mode number, $n$ is the refractive index of the core and $R$ being the radius of the resonator. The output of the solver gives the eigenfrequency which is used to get an idea of the actual refractive index of the medium for that particular optical frequency using the Sellmeier equation for the material of the waveguide [13]. The Sellmeier equation for the $Si_3N_4$ core is

$$n^2(\lambda) = 1 + \frac{3.0249\lambda^2}{\lambda^2 - 0.1353406^2} + \frac{40314\lambda^2}{\lambda^2 - 1239.842^2}, \tag{3}$$

where $\lambda$ is in units of micrometer. This iterative process continues until the difference between the resonance frequencies obtained from two consecutive iterations lies within a certain threshold (T). With an increase in the size of the resonator, the number of modes within a fixed spectral range increase, thereby increasing the simulation time. Machine learning significantly speeds up the dispersion optimization and is independent of the resonator size. From the resonance frequencies obtained from the simulation, the dispersion parameter $D_{int}$ is calculated. In this work, we use $Si_3N_4$ as resonator material. Introducing an ML model requires a dataset for the purpose of training and testing. For the generation of the dataset for the $Si_3N_4$ ring resonators, the radii of the cores are varied from 30 μm to 130 μm at an interval of 20 μm, whereas, their widths are chosen from 1 μm to 2 μm in steps of 0.1 μm. Similarly, equally spaced height values are considered from 500 nm to 800 nm in steps of 50 nm. The cladding considered here is made of fused silica and has a width and height of 4 μm each. The simulated dataset of 462 samples contains resonance wavelength in the range between 1 μm and 2 μm. For the experimental data, the quadratic fit of the $D_{int}$ is used as input vector of the predictor to retrieve the three waveguide parameters as output.

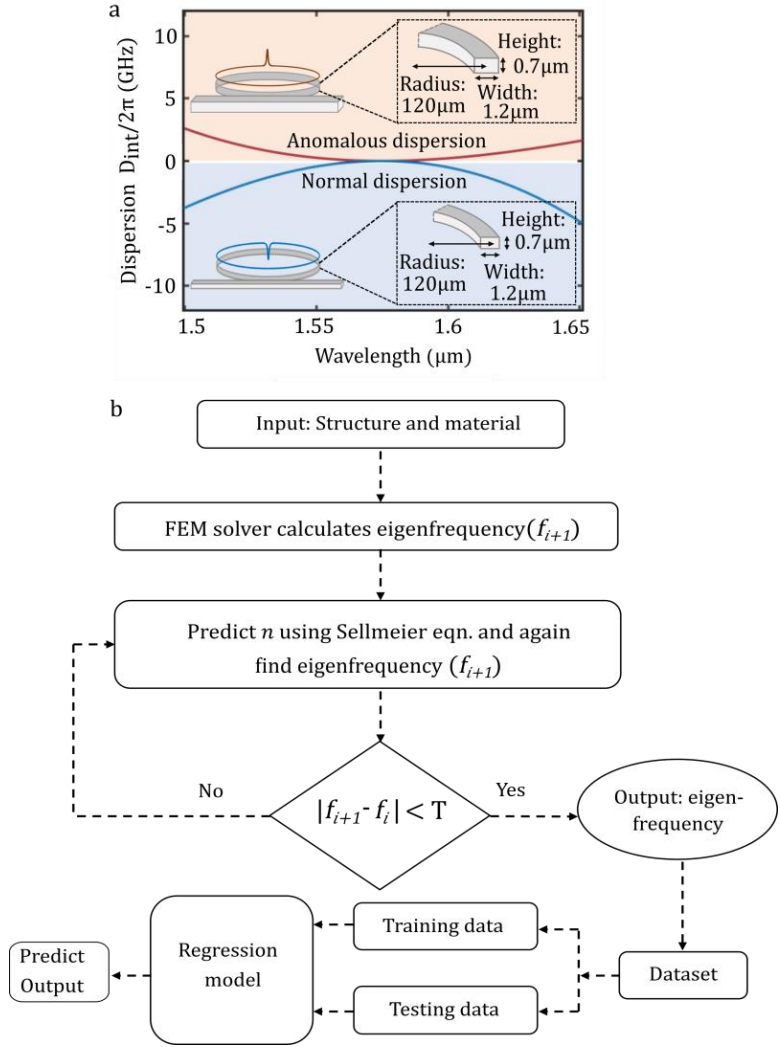

Fig. 1. (a) Influence of microresonator geometry on the integrated dispersion. Inset: The structural dimensions of the resonator core cross-section are shown. (b) Flowchart for simulating the resonance frequencies from a given structure and material. After completion of this iterative process, the data is used to calculate the dispersion parameter. Part of the dataset is used to train the regression model and the rest is used to test the prediction of the model.

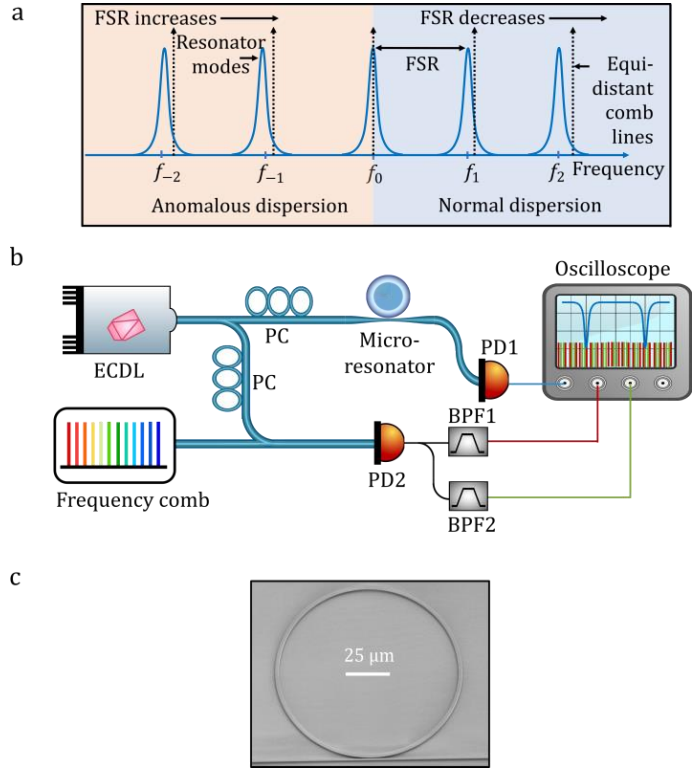

Fig. 2. (a) Variation of FSR with frequency for different kinds of dispersion. (b) The experimental setup used to measure the dispersion for the various microresonators. ECDL: External cavity diode laser, PC: Polarization controller, PD: photodiode, BPF: Bandpass filter. (c) Scanning Electron Microscope (SEM) image of the 50-μm-radius resonator.

The variation of FSR over frequency is illustrated in Fig. 2(a) for both normal and anomalous dispersion regime. To verify the regression model, trained from the simulation dataset, it is important to check whether comparable structural details are acquired when dispersion parameters obtained from experiments are fed as input. Fig. 2(b) shows the experimental setup to collect the transmission spectrum as well as the calibration point for dispersion estimation [31]. The SEM image for the 50-μm-radius resonator is portrayed in Fig. 2(c). The external diode cavity laser (ECDL) is swept from 1520 nm to 1615 nm and a microresonator transmission spectrum is recorded by an oscilloscope. Simultaneously, frequency calibration points are obtained by beating the sweeping diode laser with a stable commercial frequency comb (repetition rate 100 MHz), filtered through two bandpass filters (central frequencies at 35 MHz and 45 MHz with 2% and 1% filter bandwidths respectively. The frequency calibration points are simultaneously recorded by the oscilloscope.

### 3. Results and discussions

*3.1 Simulation results*

The performance of the geometrical parameter prediction model on the simulated dataset using RF and Decision Tree (DT) regressors are presented in this section. Both RF and DT algorithms has been implemented using the "Scikit-learn" library in python. The performances are evaluated in terms of mean absolute percentage error (MAPE). It is defined as,

$$\text{MAPE} = \frac{100\%}{n}\sum_{i=1}^{n}\left|\frac{x_i^{\text{pred}} - x_i^{\text{act}}}{x_i^{\text{act}}}\right|, \quad (4)$$

where $x_i^{\text{act}}$ and $x_i^{\text{pred}}$ are target and predicted values for $i^{th}$ sample respectively. In each case, the dataset has been divided randomly into two sections, one section (75% of the total samples) is used for training the model and the rest (25% of the total samples) for testing. Both DT and RF have hyperparameters, which control the extent of nonlinearity that can be modelled by the used architecture. In a DT regression model or a regression tree, starting from the root nodes, the dataset is split at each internal node depending upon the feature values. The condition of splitting in each node is decided by minimizing the root mean square error between the samples and the node threshold. We can choose the method of splitting at each node, the maximum depth of a tree and the minimum number of samples in a leaf node. An RF model consists of many DTs. From the training dataset, a section is selected with replacement for each of the trees. For each DT, parallel data fitting occurs and, in the end, an average of all the tree predictions is used as the final output of the RF model. In this case, the number of trees, amount of data used to build each tree, etc. are the hyperparameters that can be tuned to achieve better performance. In this work, we tuned the regressors over many possible combinations of the hyperparameters and selected the set of values that gives the best performance, a process known as grid search. Table 1 presents the best hyperparameters for the two ML regressors and their corresponding performances.

Table 1. Performance of regressors for inverse design

| Name of Model | DT | RF | RF |
|---|---|---|---|
| Data type | Not normalized | Not normalized | Normalized |
| Best model Hyper-parameters | Max. depth: 12<br>Min. samples leaf: 7 | Max. depth: 20<br>Estimators: 200 | Max. depth: 80<br>Estimators: 180 |
| MAPE (radius pred.) | 27.16 | 12.52 | 21.43 |
| MAPE (width pred.) | 9.15 | 3.14 | 1.89 |
| MAPE (height pred.) | 22.68 | 15.01 | 6.34 |

It can be seen that the DT model yields a consistent yet poor prediction accuracy. However, when using RF, the performance improves drastically. The optimal model of RF regressor has 200 different DTs. It can be seen that the MAPE for all the structural parameters are less than or around 15% for the worst case. Normalization of the dataset affects the performance of ML algorithms. Therefore, we performed the training step for RF regressor again with max-min normalized data and the results are presented in Table 1. From a comparison of the columns in Table 1, it can be seen that normalization of input and output dataset deteriorates the performance of the RF model. One of the main concerns for our model is that overfitting may affect the performance of the model due to the small size of the training data. Overfitting is encountered when a regressor increases nonlinearity to a great extent such that it fits to all of the datapoints in the training dataset, thereby reducing the performance on the test dataset. We followed cross-validation to tackle this problem. 4-fold cross-validation has been used for all of the above regressors. In this process, the total training dataset is divided into four parts in each training step, the training is done with three parts and the model is validated with the remaining section. Fig. 3 shows a histogram that illustrates the overall performance of the prediction model on the simulated data. The black dashed line shows that the error (MAPE) is less than 10% for most of the samples for all the structural parameters. In the inset, it can be seen that with increasing number of trees in an RF model, the negative mean absolute error (NMAE) decreases initially, but saturates afterwards. The performance of our model has motivated us to apply the ML model, trained on simulated data, to predict structural parameters from experimental data. In order to check the effect of the prediction error (on average 10%) on the resulting dispersion we calculated the change in dispersion induced by increasing each structural parameter (radius, height, width) separately by 10%. The corresponding MAPE for the integrated dispersion across 1500 nm to 1600 nm (experimental range) are found to be 0.01%, 0.04% and 0.05% respectively, which is not significant for these small geometry prediction errors.

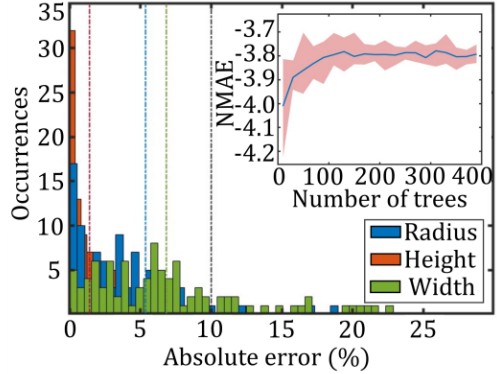

Fig. 3. Histogram showing the prediction performance of the optimized RF model on simulated data. The dashed lines indicate the median values for each parameter. Inset: Negative mean absolute error of the RF model versus number of trees in the model.

*3.2 Experimental results*

To validate the accuracy of our prediction model, we fabricated two $Si_3N_4$ ring microresonators, and measured their dispersion, then fed the dispersion values to the RF model and finally, compared the predicted structural details with the original designs. The resonators under test have different radii and waveguide heights but same waveguide width. The transmission profiles of these two resonators with radii of 50 μm and 100 μm are shown in Fig. 4(a) and 4(d) respectively, along with their corresponding $D_{int}$ plots in Fig. 4(b) and 4(e). The pump wavelength is considered to be around 1557 nm. Since the transmission plots contain multiple mode families, the one with higher Q-factor has been used for dispersion measurement. The resonators have loaded Q-factors around $5 \times 10^5$ (around 0.3 GHz linewidth). The dimensions of the actual and the predicted values for both the experiments have been listed in Fig. 4(c) and (f). One observation that can be made from the simulations is that $D_{int}$ is highly influenced by width and height of the resonator as they determine whether it will lie in the normal or anomalous regime (c.f. Fig. 1a). There are fabrication uncertainties regarding the accuracy of the structural parameters (around 10 nm for the best case) and also there can be additional impurities or inclinations of the core walls, which alter the dispersion curve and can contribute to the deviation between the actual and predicted geometrical parameters. Moreover, the avoided mode crossing (AMX) [32], as seen prominently in Fig. 4(e) around 1560 nm, changes slightly the quadratic fit of the dispersion profile. The highly sensitive ML models are influenced by such minute deviations. These kinds of crossings are not encountered during the simulations as the walls of the core are considered to be exactly perpendicular which is impossible to maintain during fabrication of the resonators. The effect of these crossings can be negated if the FSR is small or the scanning range is broader so that we can have more and more modes and thus the fitting becomes easier. Most importantly, if the dataset can be increased along with decreasing the intervals between the chosen structural parameters, the prediction ML model can be made better.

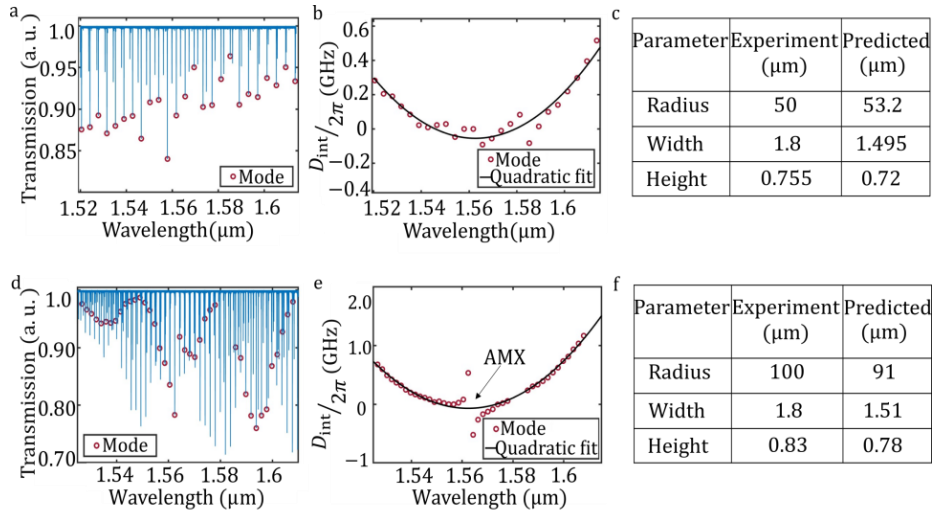

Fig. 4. Microresonator resonance spectra and corresponding dispersions used to predict the structural parameters of the microresonator. (a) Transmission spectrum of a 50-μm-radius $Si_3N_4$ resonator with its corresponding fundamental mode family being highlighted. (b) Integrated dispersion for the fundamental mode along with a quadratic fit for the spectrum in panel (a). (d), (e) The spectrum and dispersion for a 100-μm-radius resonator. (c), (f) Dimensions of the resonators predicted by the RF model from the experimentally obtained integrated dispersion values and comparison to the actual structure.

## 4. Conclusion and Outlook

We have trained a machine learning model on simulated data for predicting the structure of integrated ring resonators from their dispersion profiles generated by varying the height, radius and width of microresonators in finite element simulations. Comparing "decision tree" and "random forest" algorithms, we concluded that "random forest" yields better performance for structure prediction. The optimized and tuned model has been used for structure prediction based on experimentally measured dispersion profiles from silicon nitride resonators. Increasing the size and spectral width of the training data set could further increase the accuracy of the structure prediction. With the ever-growing possibilities in fabrication of photonic integrated circuits, machine learning can play an important role to precisely engineer dispersion for a large number of different applications. In particular, this enables the generation of spectrally tailored optical soliton frequency combs in chip-integrated microresonators. In the future, this work can be extended to more complex geometries with organically changing resonator geometries, which enables even more broadband control of microresonator mode structures.

**Funding.** European Union's H2020 ERC Starting Grant "CounterLight" 756966; Marie Curie Innovative Training Network "Microcombs" 812818; and the Max Planck 329 Society.

**Disclosures.** The authors declare no conflicts of interest.

**Data Availability.** Data from this article is available from the authors upon reasonable request.